\documentclass{article}

\usepackage[final]{css}

\usepackage[utf8]{inputenc} 
\usepackage[T1]{fontenc}    
\usepackage{hyperref}       
\usepackage{url}            
\usepackage{booktabs}       
\usepackage{amsfonts}       
\usepackage{nicefrac}       
\usepackage{microtype}      
\usepackage{amsmath}
\usepackage{graphicx}

\title{FastThaiG2P: Lightning-fast Thai Grapheme-to-phoneme Conversion for Voice Agent Pipelines}

\author{
  Charin Polpanumas\\
  AWS\\
  \texttt{charipol@amazon.com} \\
}

\begin{document}

\maketitle

\begin{abstract}

FastThaiG2P provides sub-millisecond Thai grapheme-to-phoneme conversion for text-to-speech pipelines (International Phonetic Alphabet and Kokoro-TTS conventions) using a PyThaiNLP-tokenized, extensible dictionary and normalization rules for common Central Thai speech. The approach achieves an average latency of 0.15 ms per utterance on a benchmark of 27,242 synthetically generated utterances, of which 30\% is spent on tokenization, 12\% on normalization, and 58\% on out-of-vocabulary fallbacks (0.5\% OOV rate). To demonstrate its effectiveness, we used FastThaiG2P to phonemize Som-TTS, an open dataset containing 20 hours of grapheme-and-audio pairs, then trained an 82M-parameter StyleTTS 2 model based on a Kokoro-TTS recipe. The resulting model vocalizes intelligible Thai speech suitable for prototyping and development at 0.25 real-time factor (4x real-time) with ONNX inference on CPU.

\end{abstract}

\section{Introduction}

\paragraph{Challenges in Thai Phonemization} Thai text-to-speech (TTS) systems face a unique set of challenges at the grapheme-to-phoneme (G2P) stage. Unlike languages with relatively transparent orthography, alphasyllabary Thai script exhibits several properties that complicate phonemization: 1) words and sentences are written without explicit boundaries, requiring segmentation as a prerequisite to phonemization 2) Thai is a tonal language with five lexical tones whose surface realization depends on syllable structure, consonant class, vowel length, and final consonant 3) vowels are written as discontinuous graphemes that surround their onset consonant and some are polyfunctional, serving as standalone vowels as well as components of other vowels 4) multiple classes of non-transparent grapheme-to-phoneme mappings--namely leading consonant tone shift, reduced vowels absent from the orthography, context-dependent ligatures, false clusters, and homographs--mean that character-level rules alone are insufficient 5) real-world Thai text contains a dense mixture of numbers, abbreviations, symbols, English loanwords, and code-switched tokens that must be verbalized before phonemization. For real-time voice agent pipelines such as call center agents, conversational AI assistants, and live captioning systems, the G2P stage must be both comprehensive and fast. A latency budget of 500–1,000 ms for the entire TTS pipeline leaves little room for slow preprocessing.

\paragraph{Existing Approaches} Several tools provide Thai G2P functionality, spanning rule-based, dictionary-based, and model-based approaches. \textbf{TLTK} \cite{aroonmanakun2018tltk} performs syllable-level G2P via a trigram-based syllable segmenter and rule-based phonological mapping. It handles regular Thai orthography well but relies entirely on rules without a pre-built dictionary, limiting accuracy on irregular words and loanwords. Its runtime recompiles regex patterns on every call, resulting in high latency for batch processing, up to 2 ms per utterance on average over our 27,242-utterance synthetic dataset. \textbf{Epitran} \cite{mortensen2018epitran} offers rule-based grapheme-to-IPA mapping for 61 languages including Thai via a character map with pre- and post-processing rewrite rules. Its Thai mode reorders leading vowels, applies coda neutralization, and maps consonants/vowels to IPA segments. However, it explicitly discards tone marks without inferring tonal values from syllable structure, requires pre-segmented word input, and does not include word segmentation or text normalization. \textbf{thai-g2p} \cite{phatthiyaphaibun2020thaig2p} uses a seq2seq, MarianMT-based model \cite{junczys2018marian} trained on Wiktionary data to predict phonemes from Thai words. It requires external word segmentation and does not include text normalization. \textbf{CharsiuG2P} \cite{zhu2022charsiu-g2p} is a ByT5-based \cite{xue2022byt5} multilingual neural G2P covering 100 languages including Thai. It achieves a phoneme error rate (PER) of 26.9\% and word error rate (WER) of 59\% and produces IPA with tonal notation for Thai. However, it requires pre-tokenized word input, has higher inference latency than dictionary lookup due to autoregressive decoding, and does not include text normalization. None of these tools simultaneously provide 1) a large curated IPA dictionary for Thai, 2) comprehensive text normalization for spoken forms, 3) sub-millisecond per-utterance latency, and 4) robust OOV handling, all of which are pre-requisites for inputs to phoneme-based TTS architectures \cite{hansen2023piper} \cite{hexgrad2024kokoro_hf}.

\paragraph{Phonemization-free Thai TTS} Recent grapheme-based TTS models can synthesize Thai speech directly from text without explicit G2P. MMS-TTS \cite{pratap2024scaling} provides a lightweight VITS \cite{kim2021conditional} model for Thai but suffers from high error rates (CER 18.28\% on Common Voice 13 Thai test split \cite{ardila2020common}). ThonburianTTS \cite{aung2025thonburiantts} finetunes F5-TTS \cite{chen2025f5} on 969 hours of Thai speech, achieving CER 8.70\% on Common Voice 13 Thai test split. JaiTTS \cite{karnjanaekarin2026jaitts} adapts VoxCPM-0.5B \cite{zhou2025voxcpm} for Thai with CER of 1.94\% on their internal benchmark. Multilingual architectures including Qwen3-TTS \cite{hu2026qwen3}, OmniVoice \cite{zhu2026omnivoice}, VoxCPM-0.5B \cite{zhou2025voxcpm} and VoxCPM2 \cite{zhou2026voxcpm2} support Thai among many languages. However, these models require 336M–2B parameters and GPU inference. For deployment scenarios requiring CPU-only inference, sub-second latency, or minimal memory footprint such as edge devices and cost-sensitive batch pipelines, a phoneme-based approach with a compact acoustic model \cite{kim2021conditional} \cite{hansen2023piper} \cite{li2023styletts} \cite{hexgrad2024kokoro_hf} remains the practical choice thus the necessity of a fast and accurate phonemization front-end.

\paragraph{FastThaiG2P Enables Intelligible, Low-Latency, CPU-based TTS} This paper presents FastThaiG2P, an open-source Thai G2P library that bridges the gap between dictionary-based accuracy and the speed required for real-time phoneme-based TTS. The system combines a 62,112-word IPA dictionary with comprehensive text normalization and a rule-based out-of-vocabulary (OOV) fallback, achieving 0.15 ms per utterance on a 27,242-utterance benchmark. Paired with a Thai-finetuned Kokoro-82M \cite{hexgrad2024kokoro_hf} checkpoint based on StyleTTS 2 \cite{li2023styletts}, the full pipeline produces intelligible Thai speech at 0.25 real-time factor (RTF) on CPU with a total footprint of 330 MB. \textbf{Our contributions are}: 1) a 62,112-word IPA dictionary assembled from Wiktionary, LLM-generated transcriptions validated against phonological rules, and manual overrides 2) a text normalization pipeline covering 15 categories of non-speakable text including numbers, Thai abbreviations, units, symbols, phone numbers, emails, and time patterns 3) up to 15x latency reduction over the TLTK baseline through regex caching and import-time initialization 4) an end-to-end TTS demonstration using FastThaiG2P to train a Thai Kokoro-82M model achieving 0.25 RTF on CPU. The library is released under Apache-2.0 at \texttt{github.com/aws/FastThaiG2P}.

\section{FastThaiG2P System Design}

FastThaiG2P implements a four-stage pipeline, sequentially, text normalization, tokenization, phoneme dictionary lookup, and fallback G2P.

\subsection{Text Normalization} 

The normalizer converts non-vocalizable scripts to vocalizable ones before tokenization. This acts as the deterministic, last line of defense against invalid TTS inputs in case prompting the large language model (LLM) fails in a cascading voice agent pipeline (automatic speech recognition (ASR) → LLM → TTS). Processing occurs in a fixed order designed to prevent ambiguity:

\begin{itemize}
    \item Expand maiyamok, the Thai word repetition marker
    \item Convert Thai numerals to Arabic digits
    \item Read email addresses
    \item Read English abbreviations and brand names (transliteration dictionary)
    \item Read units (kg, km, °C, etc.)
    \item Read symbols (\%, +, ×, etc.)
    \item Read time patterns (14:30, 23:12, etc.)
    \item Read phone numbers (digit-by-digit per group)
    \item Read alphanumeric identifiers (ORD-001)
    \item Read comma-separated numbers (1,000)
    \item Read decimal and plain numbers
    \item Read Thai abbreviations
    \item Read residual Latin characters
\end{itemize}

Short numbers (6 digits or shorter) use Thai place-value reading while long numbers (7 digits or longer) are read digit-by-digit, matching the Thai convention for phone numbers, account numbers, and IDs.

\subsection{Tokenization} 

After normalization, the text is segmented into words using PyThaiNLP's `newmm` (maximum matching) engine \cite{phatthiyaphaibun2023pythainlp} with a custom dictionary. The dictionary (`data/dict.txt` and `data/ipa.json`) serves both as a custom word-list for the segmentation and lookup keys for corresponding IPA entries. This coupling ensures every token produced by the word segmenter has an IPA barring OOV tokens.

\subsection{Dictionary Construction}

The 62,112-entry IPA dictionary was built from three main sources, from highest to lowest merge priorities:

\paragraph{Manual Overrides} (\texttt{sources/manual\_overrides.json}) manually verified IPA transcriptions for words where we could not find references from Wikitionary \cite{ylonen2022wiktextract} and/or automated methods fail. This serves as a human-in-the-loop lever for continuous dictionary improvement. 

\paragraph{Wiktionary} (\texttt{sources/wiktionary\_ipa.json}) about 13,000 entries extracted from the Thai Wiktionary dump. These follow the English Wiktionary IPA convention for Thai, which served as the reference standard for our transcription format.

\begin{figure}[h]
    \centering
    \includegraphics[width=1\textwidth]{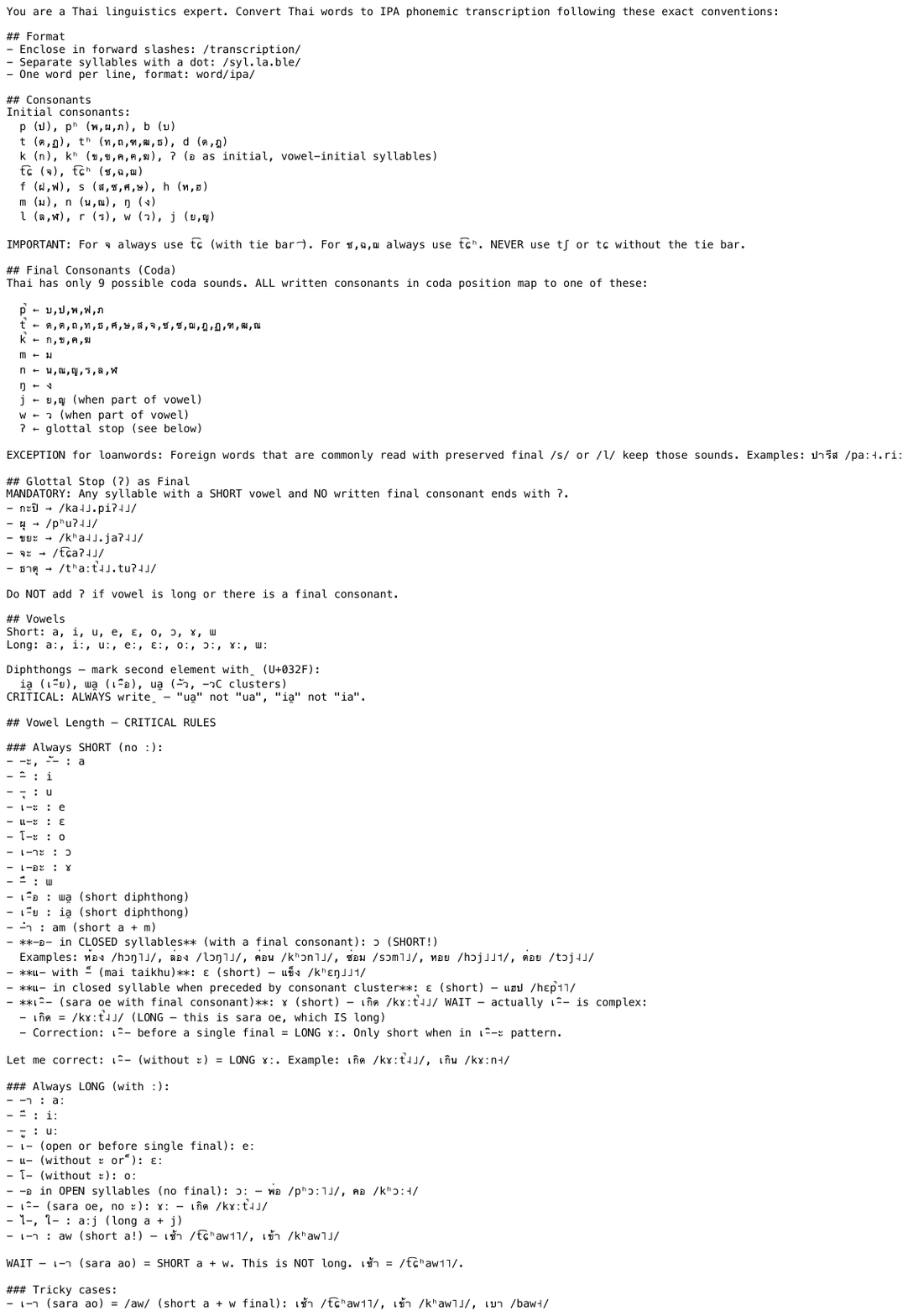} 
    \caption{Prompt for IPA Transcription}
    \label{fig:prompt_ipa}
\end{figure}

\paragraph{LLM Transcription} (\texttt{sources/generated\_ipa.json}) approximately 49,000 entries generated by Claude Opus 4.6 via Amazon Bedrock. Generation was performed in batches of 100 words using a detailed prompt (Figure \ref{fig:prompt_ipa}; \texttt{scripts/ipa\_prompt.txt}) specifying the IPA convention, phonological constraints, and example transcriptions. Each response was validated by checking that every character falls within a 38-codepoint phoneme inventory whitelist (consonants, vowels, Chao tone letters, combining diacritics, and separators); entries with out-of-inventory characters or missing slash delimiters were rejected and logged to \texttt{generated\_ipa\_invalid.json}.

Before batch generation, the prompt was validated against 500 random Wiktionary entries as a validation set (\texttt{scripts/validate\_prompt.py}). With TTS as the objective, errors were classified as non-fatal (tone mismatch, vowel length, diphthong marker, coda variant) or fatal (wrong syllable count, wrong initial consonant or vowel quality). Pali/Sanskrit loanwords with irregular readings were excluded from the fatal count since Wiktionary handles them directly. The final prompt has 78.8\% exact match rate and 1.8\% fatal error rate on the validation set. 

\subsection{OOV Fallback} 

Words not found in the dictionary receive IPA transcription via a rule-based fallback vendored from TLTK \cite{aroonmanakun2018tltk}. The fallback segments the word into syllables using trigram statistics (`data/fallback/sylseg.3g`), maps each syllable to a romanized pronunciation using consonant class, vowel pattern, and tone rules, then converts the romanization to our IPA convention (Chao tone letters, aspiration, unreleased stops). The fallback always produces phonologically valid Thai IPA but may not match the conventional pronunciation of irregular words (loanwords, Pali/Sanskrit compounds with silent letters). Frequently encountered OOV words should be manually added to the dictionary.

\paragraph{Regex Cache Optimization} The original TLTK implementation recompiles regular expressions on every function call. The \texttt{sylparse} function iterates over hundreds of syllable pattern regexes, calling \texttt{re.match(pattern\_string, ...)} at every character position of the input word. Python's internal regex cache, a Least Recently Used (LRU) cache limited to 512 entries, overflows when the number of unique patterns exceeds this limit, causing repeated recompilation. In the vendored version, we introduced an explicit unbounded regex cache at all three hot loops in the fallback code: syllable parsing, syllable enumeration, and rule loading. Combined with pre-loading the syllable rules at import time rather than on first call, this reduced the overall G2P latency from approximately 2 ms per utterance to 0.15 ms per utterance, up to 15x improvement. The majority of the savings come from eliminating redundant regex compilation in the fallback path, which previously dominated wall-clock time even though it was invoked on a minority of tokens.

\subsection{IPA Convention}

\begin{figure}[!h]
    \centering
    \includegraphics[width=0.8\textwidth]{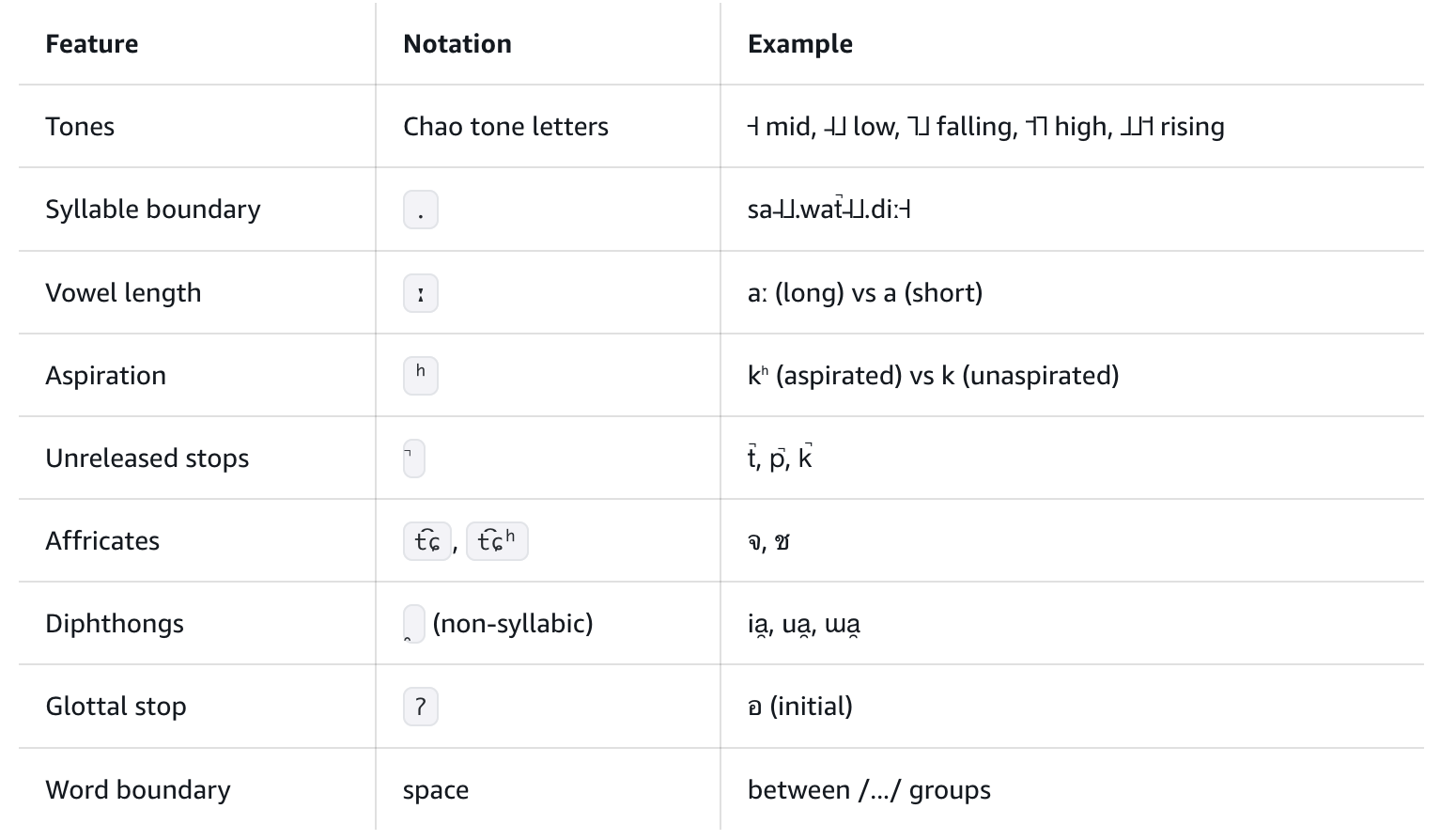} 
    \caption{IPA Convention}
    \label{fig:ipa_convention}
\end{figure}

We follow the English Wiktionary Thai IPA transcription standard \cite{wiki:mos_pronunciation_ipa} using Chao tone letters (Figure \ref{fig:ipa_convention}). The inventory uses 38 phoneme-related codepoints. 

\begin{figure}[!h]
    \centering
    \includegraphics[width=0.8\textwidth]{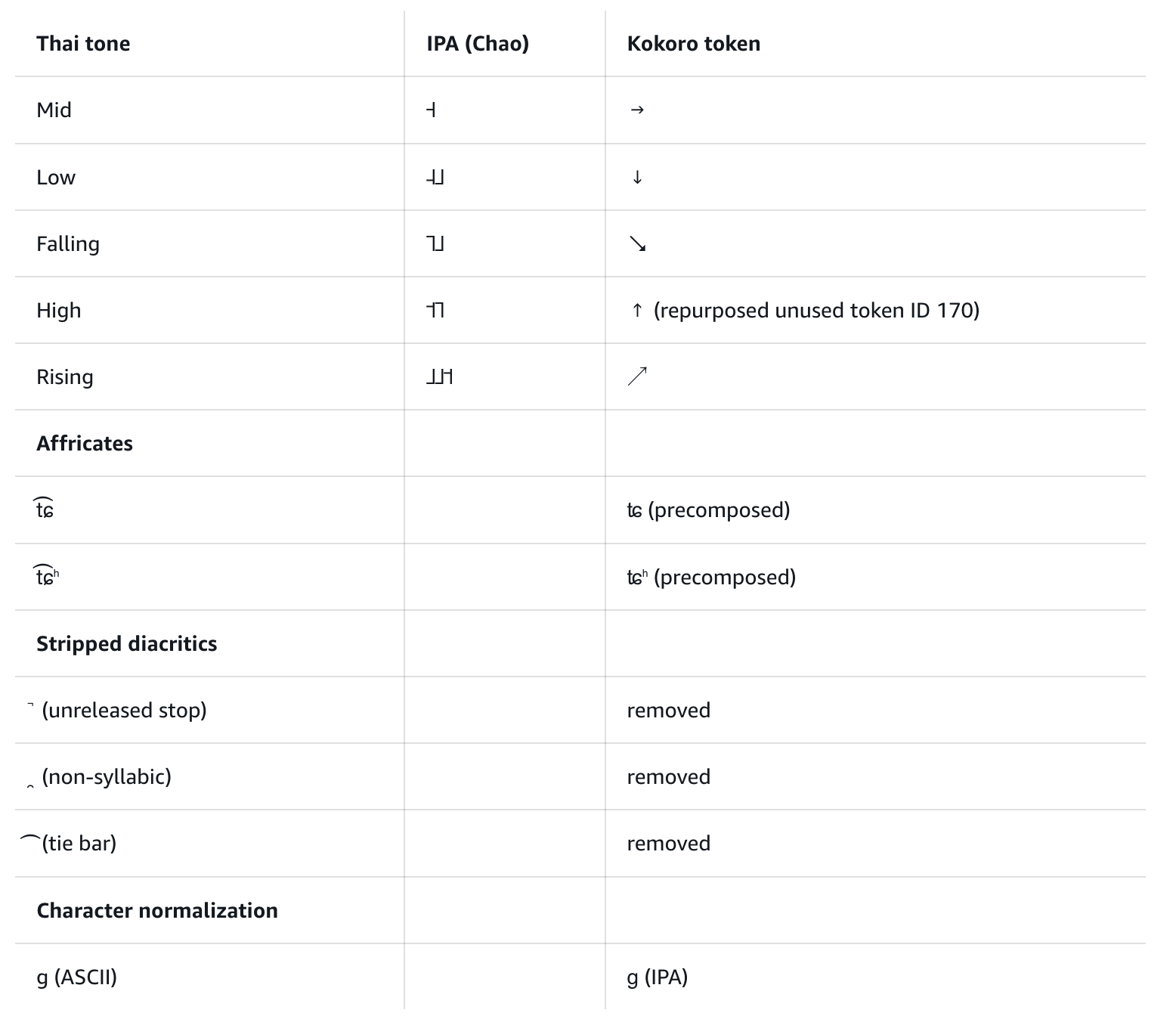} 
    \caption{IPA-to-Kokoro Mapping}
    \label{fig:ipa_kokoro_mapping}
\end{figure}

\paragraph{IPA-to-Kokoro Mapping} Kokoro-82M was pretrained on English and does not natively support Thai. Its phoneme vocabulary includes four intonation markers used for English prosody; however Thai requires five tonally distinct markers. A naive mapping would force a merger between high and rising tones, which are phonemically contrastive. We repurposed an unused token ID 170 as the high tone marker, yielding the following five-tone mapping. We also stripped diacritics that only exist in IPA as well as substitute affricates and \texttt{g} character with the corresponding Kokoro equivalents (See Figure \ref{fig:ipa_kokoro_mapping}).

\section{Latency Profiling}

\begin{figure}[!h]
    \centering
    \includegraphics[width=0.8\textwidth]{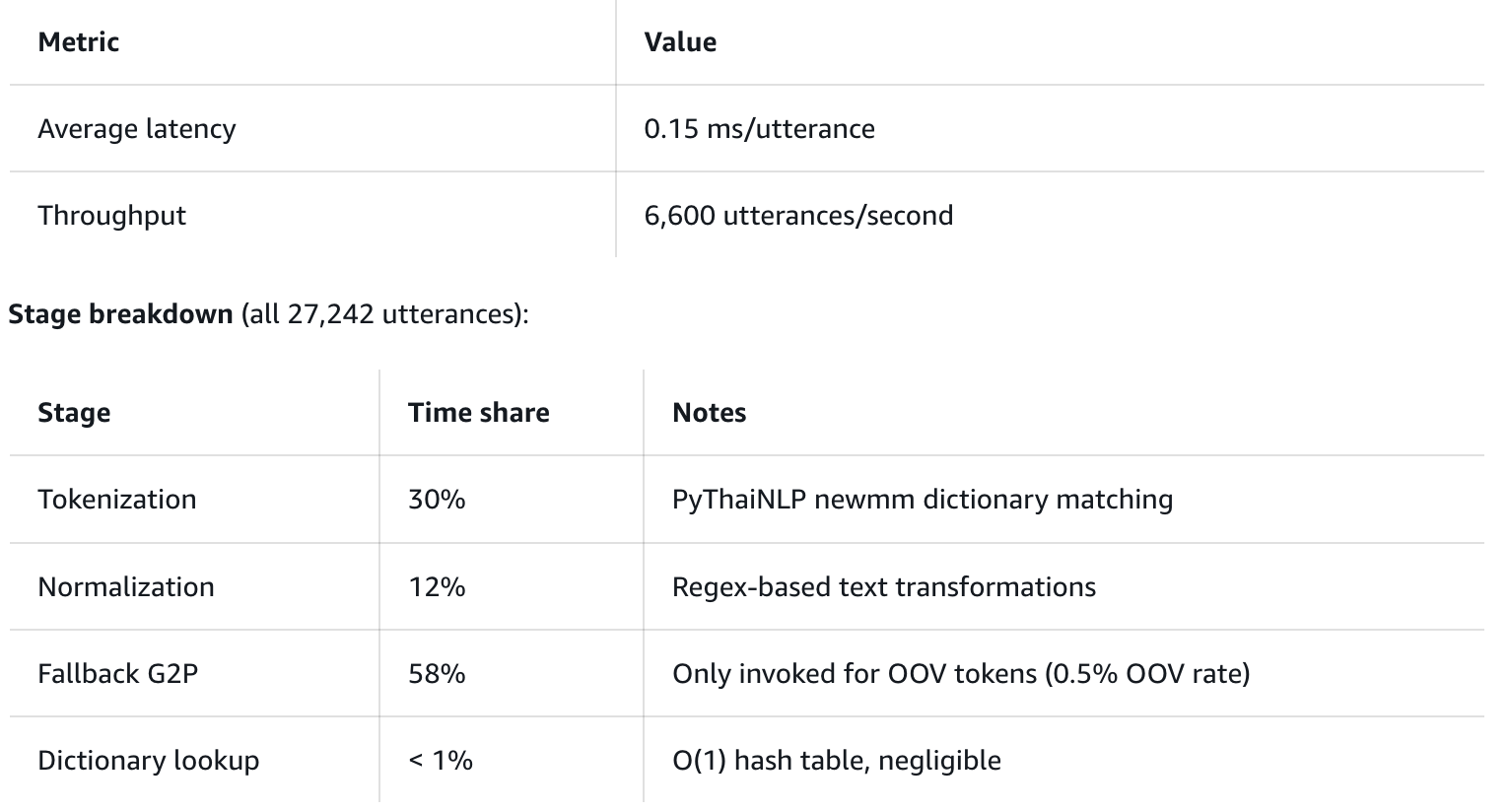} 
    \caption{Latency Profiling Result}
    \label{fig:latency_result}
\end{figure}

Latency was measured on a corpus of 27,242 synthetically generated Thai utterances (\texttt{data/synthetic/utterances.jsonl}) representing conversational text, numbers, abbreviations, and domain-specific vocabulary typical of voice agent interactions. Profiling was performed using \texttt{scripts/profile\_g2p.py} which measures end-to-end latency across all utterances and per-component timing with cProfile instrumentation on a single-threaded CPU setup with Python 3.11. Despite only 0.5\% of tokens being OOV, the fallback dominates wall-clock time (58\%) because trigram-based syllable parsing is orders of magnitude more expensive per-token than a hash lookup. The pipeline exhibits predictable latency with no neural network inference, no disk input-output per call, and no per-call memory allocation. See Figure \ref{fig:latency_result}.

\section{Text-to-Speech Demonstration}

To validate that FastThaiG2P enables a minimally viable Thai TTS system with CPU-only inference, we trained a Thai-finetuned Kokoro-82M checkpoint end-to-end. We used Kokoro-82M \cite{hexgrad2024kokoro_hf}, an 82-million-parameter architecture based on StyleTTS 2 \cite{li2023styletts}. The model was Thai-finetuned using the kikiri-tts training recipe \cite{semidark2026kikiritts}, which implements a two-stage training process: stage 1 for text-to-mel alignment and duration prediction, and stage 2 for adversarial training with a multi-scale discriminator. The model takes Kokoro-format phoneme sequences as input and produces 24 kHz audio. A voicepack vector conditions the model on a target speaker identity. We used Som-TTS \cite{pythainlp2026somtts}, an open Thai TTS dataset containing approximately 20 hours of single-speaker recordings with aligned Thai text transcripts. FastThaiG2P was used to phonemize all transcripts, converting Thai graphemes to the Kokoro phoneme format via the IPA intermediate representation. For deployment without PyTorch, the trained checkpoint was exported to ONNX format, outputting raw audio samples plus frame durations for boundary token trimming.

\begin{figure}[!h]
    \centering
    \includegraphics[width=0.8\textwidth]{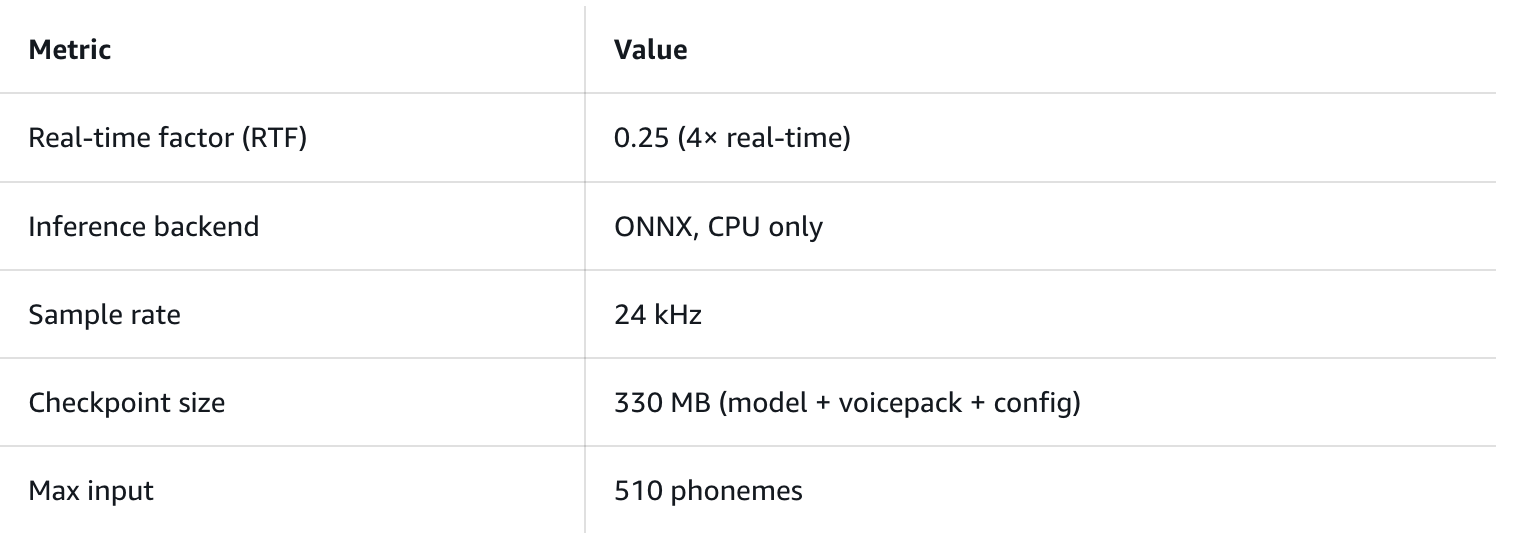} 
    \caption{TTS Inference Result}
    \label{fig:tts_result}
\end{figure}

The resulting model produces intelligible Thai speech with recognizable tonal patterns and natural rhythm for short to medium utterances at RTF 0.25 (4x real-time) on a single-threaded CPU inference. The total memory footprint is about 330 MB (See Figure \ref{fig:tts_result}). Audio quality is suitable for prototyping and development. The TTS module includes a post-processing step that removes static/noise generated during the beginning-of-sentence (BOS) and end-of-sentence (EOS) pad token windows using voicing detection to distinguish genuine speech onset from artifacts in the boundary regions.

\section{Discussion and Future Work}

\paragraph{Dictionary Coverage} The 62k-word dictionary covers standard Central Thai well but has gaps in regional vocabulary, recent loanwords, slang, and code-switched Thai-English text. We expect to periodically update the dictionary and allows seamless manual overrides for each use case.

\paragraph{Fallback Quality} The rule-based fallback produces phonologically valid IPA but may err on irregular words (silent letters in Pali/Sanskrit compounds, non-standard loanword pronunciations). A hybrid approach combining rules with a small neural model could improve OOV handling.

\paragraph{Tone Accuracy} Thai tone assignment from orthography is largely rule-governed but has exceptions such as tone-mark elision in certain compounds, dialectal variation. The current system follows standard Central Thai rules; regional variants are not modeled.

\paragraph{TTS Quality Evaluation} No formal MOS study has been conducted. A perceptual evaluation with native Thai speakers would quantify quality and enable comparison against commercial Thai TTS systems.

\paragraph{Future Directions} We see FastThaiG2P as an enabler for increasingly more potent phoneme-based, CPU-only TTS systems that power hybrid voice agents, agentic systems which leverage frontier models to perform complex tasks on AWS Bedrock and Sagemaker while outsourcing simpler tasks such as ASR and TTS to local devices.

\section{Appendix}\label{appendix}

\subsection{Prompt for IPA Transcription}\label{prompt_ipa}

\bibliographystyle{unsrt}  
\bibliography{references}  

\end{document}